\documentclass[11pt]{article}
\usepackage{acl2016}
\usepackage{times}
\usepackage{url}
\usepackage{latexsym}
\usepackage{paralist}
\usepackage{diagbox}
\usepackage{graphicx}

\aclfinalcopy 

\newcommand{\Expect}{{\rm I\kern-.3em E}}

\title{Towards Multi-Agent Communication-Based Language Learning}

\author{Angeliki Lazaridou \and Nghia The Pham \and Marco Baroni\\
            University of Trento\\
            {\tt \{{angeliki.lazaridou|thenghia.pham|marco.baroni\}}@unitn.it}}

\date{}

\begin{document}
\maketitle
\begin{abstract}
  We propose an interactive multimodal framework for language
  learning. Instead of being passively exposed to large amounts of
  natural text, our learners (implemented as feed-forward neural
  networks) engage in cooperative referential games starting from a
  tabula rasa setup, and thus develop their own language from the need
  to communicate in order to succeed at the game. Preliminary
  experiments provide promising results, but also suggest that it is
  important to ensure that agents trained in this way do not develop
  an ad-hoc communication code only effective for the game they are
  playing.
\end{abstract}

\section{Introduction}
\label{sec:intro}

One of the most ambitious goals of AI is to develop intelligent
conversational agents able to communicate with humans and assist them
in their tasks. Thus, communication and interaction should be at the
core of the learning process of these agents; failure to integrate
communication as their main building block raises concerns regarding
their usability. However, traditional machine-learning approaches to
language are based on static, passive, and mainly supervised regimes
(e.g., as in applications to parsing, machine translation, natural
language generation).  Computational ``passive learners'' receive a
lot of annotated data and, by observing regularities in them, discover
patterns they can apply to new data.  While this is a great way to
learn general statistical associations, it is very far from
interactive communication, which proceeds by an active and incremental
updating of the speakers' knowledge states.


In the language community, after the seminal work on the
``blocks-world'' environment of \newcite{Winograd:1971}, we are now
experiencing a revival of interest in language learning frameworks
that are centered around communication and interaction (e.g., the
\emph{Roadmap} of \newcite{Mikolov:etal:2015a} or the more recent
\emph{dialogue-based learning} proposal of \newcite{Weston:2016}).
Similar trends are also taking place in other fields of Artificial
Intelligence, witness the revival of interest in \emph{game playing}
with the recent ground-breaking results of DQN on
Atari~\cite{Mnih:etal:2015} and AlphaGo~\cite{Silver:etal:2016},
following precursors such as DeepBlue~\cite{Campbell:etal:2002} and
TD-Gammon~\cite{Tesauro:1995}.

Focusing on language, current approaches to communication-based
language learning simulate interactive environments in diverse ways,
e.g., by having agents interacting directly with humans or other
scripted agents.  Both approaches exhibit potentially important
limitations.  The \emph{human-in-the-loop} approach (e.g., the SHRLDU
 program of Terry Winograd, robots learning via interacting
with humans as in \newcite{Tellex:etal:2014}) faces serious
scalability issues, as active human intervention is obviously required
at each step of training. Scripted Wizard-of-Oz
environments~\cite{Mikolov:etal:2015a} shift the burden of heavy manual
engineering from the learning agent to designing the right behaviour
for the programmed teaching agents.

In this work, we are proposing a radically different research program, namely
\emph{multi-agent communication-based language learning
within a multimodal environment}. %
The
essence of this proposal is to let computational agents co-exist, so that their
co-existence constitutes the interactive environment. In this multi-agent
environment, agents need to collaborate to perform a task, and we hypothesize
that (with the right priors and constraints) developing language production and
understanding will be prerequisites to successful communication. 
Note that we are not suggesting that the ``passive'' setup
should be abandoned, as, even in the interactive paradigm, large-scale
statistical learning is expected to be important, e.g., to let agents discover
how to produce grammatical sentences, how to recognize object categories in
general or even how to provide generic descriptions of what is present in a
scene.  Still, interaction is required for many other tasks, that only make
sense within a communicative setup, e.g., how to refer to specific things, or
how to ask a good question or respond to it. 

Several points make this proposal attractive. First of all, this
framework requires minimum human intervention for designing agents,
the environment and its physics, e.g, rewards, although humans do
still need to specify the nature of the tasks that agents need to
perform.  Computational agents will co-exist (co-operate or
antagonize) and self-organize freely, interacting with each other and
being encouraged to learn in order to achieve communication. For
example, imagine the simple case in which an agent needs to have some
object that some other agent possesses, and she starts asking for it
in various ways. Only when she manages to make herself understood she
will be able to get hold of that object. The sort of learning taking
place in such setup will have to be based on \emph{active} request for
information, and it will probably foster incremental agreement by
\emph{interaction}.

\begin{figure}[t]
\center
\includegraphics[scale=0.60]{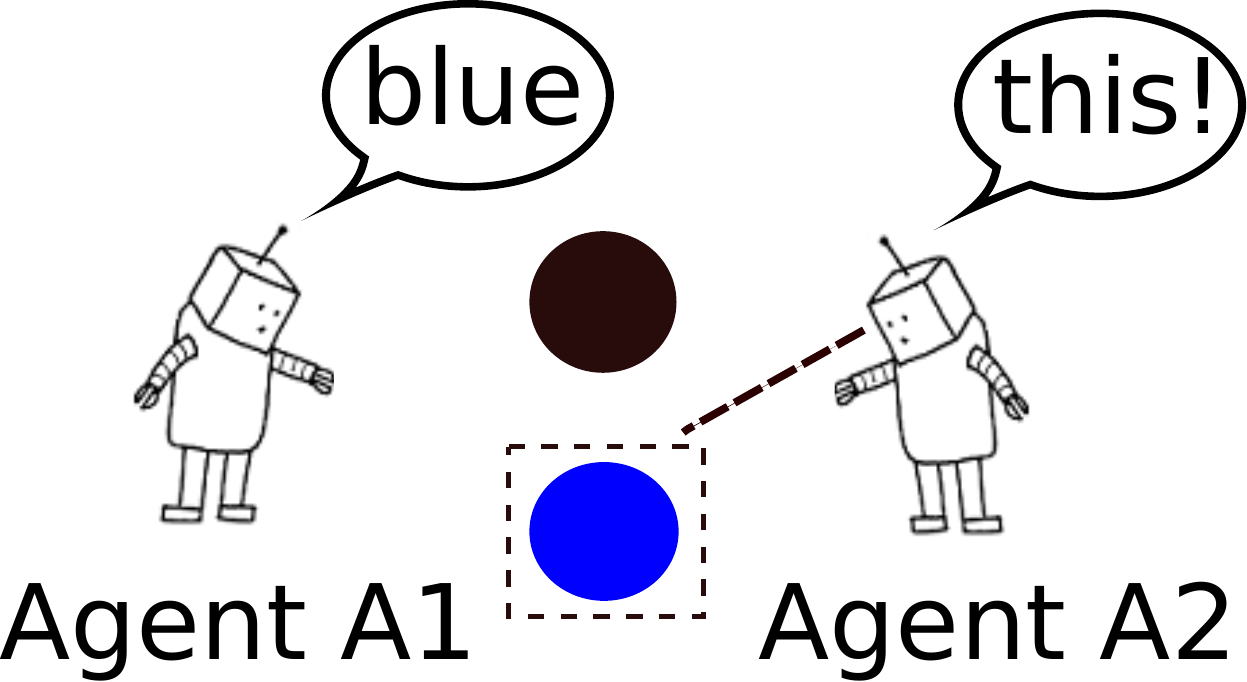}
\caption{Schematic representation of the referential game. Agent A1 has to describe
the dashed object with an attribute. Agent A2, has to guess which object A1 was referring to.
In this   example, the agents agree on the referent, so the communicative act
  was successful.}
\label{fig:vis_summary}
\end{figure}

We start by considering the most basic act of communication, namely
referring to things. We design multimodal riddles in the form of
\emph{referential games} (see Figure
\ref{fig:vis_summary})~\cite{Galantucci:etal:2012}. The speaker in
this game is asked to refer to one of the visible objects by uttering
an expression. The listener, who sees the same objects but  has no
knowledge regarding which object the speaker was asked to describe,
needs to identify it based on the speaker's expression.

Importantly, the agents start in a \emph{tabula rasa} state.  They do
not possess any form of language or understanding. They have no prior
notion of the semantics of words. 
Meanings are assigned to words (that is, the arbitrary symbols used in
our initial simulations) by playing the game and are reinforced by
communication success.  Thus, agents can agree to any sort of
conceptualization and assign to any word any kind of interpretation
that help them effectively solve the tasks. This essentially aligns
with the view of \newcite{Wittgenstein:1953} that \emph{language
  meaning is derived from usage}.

We will report next a set of pilot experiments showing that, while it
is feasible, within this multi-agent environment, to learn efficient
communication protocols succeeding in the referential game, such
protocols might not necessarily be aligned with the sort of semantics
that exists in natural language.  Interestingly, in the seminal
language evolution experiment ``Talking Heads'' of
\newcite{Steels:2015}, when two robots were left free to interact with
each other, they developed an artificial language bearing little
resemblance to natural language. Thus, we anticipate that, if we want
to move forward with this research programme, grounding the agents'
communication into natural language will be crucial, since our
ultimate goal is to be able to develop agents capable of communication
with humans. We will return to this point in
Section~\ref{sec:discussion}.

\section{A two-agent referential game simulation}

\begin{table*}[t] 
\begin{center}
\begin{tabular}{ c c c c }
& \textbf{ReferIt} & \textbf{Objects} & \textbf{Shapes} \\ 
\textbf{visual scenes} & \includegraphics[scale=0.40]{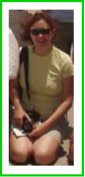}  \includegraphics[scale=0.40]{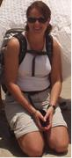}  & \includegraphics[scale=0.30]{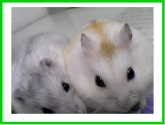}  \includegraphics[scale=0.10]{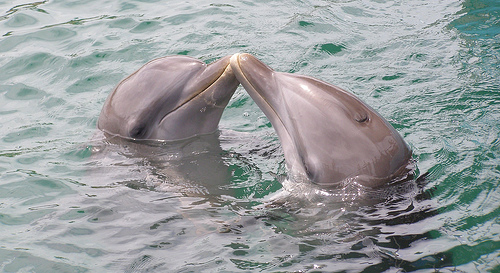}  &\includegraphics[scale=0.20]{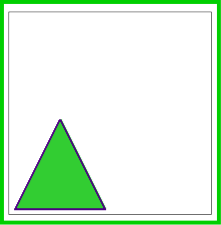}  \includegraphics[scale=0.20]{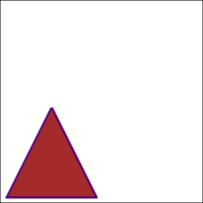} \\  
\textbf{gold attributes} & yellow & hamster & limegreen\\
\textbf{\#unique images}& 22.5k & 46k & 2.4k  \\
\textbf{\#visual scenes} &  25k & 495k &100k\\
\textbf{\#gold attributes}& 3467 & 100 & 18\\
\end{tabular}
\caption{Examples of $\langle referent, context\rangle$ pairs from our 3 datasets. Referents are marked with a green square.}
\label{fig:games}
\end{center}
\end{table*}

\subsection{The game} We propose a simple referential game
with 2 agents, \emph{A1} and \emph{A2}. The game is defined as follows:  
\begin{itemize} 
\item A1
is shown a visual scene with two objects and is told to describe the referent
with an attribute that constitutes the referring expression (\emph{RE}).
\item A2 is shown the same visual scene
without information on which is the referent, and given the RE has to ``point''
to the intended object 
\item if A2 points to the correct object, then both agents receive
a game point.  
\end{itemize} Note that this game resembles the
ReferItGame~\cite{Kazemzadeh:etal:2014} played by humans and designed to
collect RE annotations of real scenes. 

We observe that this is a \emph{co-operative} game, i.e., our agents must work
together to achieve game points. A1 should learn how to provide  accurate
phrases that discriminate the object from all others, and A2 should be good at
interpreting A1's REs in the presence of the objects, in order to point to the
correct one.  Thus, with this game A1 and A2 learn to perform \emph{referring
expression generation} and \emph{reference resolution} respectively.

\subsection{Visual Scenes} For reducing the complexity of our first
simulations, unlike the ReferItGame, we do not work with real-world visual
scenes. Instead, we construct visual scenes consisting of objects depicted in
images. Specifically, our current setup consists of visual scenes with only two
objects, the \emph{referent} and the \emph{context}.  This allows us to have
control over the complexity of the image processing required and the  game
itself, e.g., by having a referent that is visually dissimilar from the context
and can be thus easily discriminated. Towards this end, we created 3 games from
3 different datasets by controlling the type of objects involved in the visual
scene.  Each dataset focuses on some particular aspect of the referential game.

While our framework does not require obtaining gold annotations for the
RE of the referent, we apply a number of heuristics to
annotate each $\langle$referent, context$\rangle$ pair with \emph{gold
attributes} acting as REs. This will allow us to conduct
various analyses regarding the nature of the semantics that the agents assign to
the induced attributes. Table ~\ref{fig:games} exemplifies an instance of the
game for the different scenarios, and reports descriptive statistics for the 3
datasets.\footnote{Our datasets will be made available.}

\paragraph{ReferIt} The first game scenario uses data derived from the
ReferItGame~\cite{Kazemzadeh:etal:2014}. In its general format, ReferItGame
contains annotations of bounding boxes in real images with referring
expressions produced by humans when playing the game. In order to create
plausible visual situations consisting of two objects only, we synthesize scenes
by pairing each referent (as denoted by a  bounding box in the image) with a
distractor context that comes from the \emph{same} image (i.e., some other
bounding box in the image).  Each bounding box in the initial RefetItGame is
associated with a RE, which we pre-process to eliminate stop
words, punctuation and spatial information, deriving single words attributes.
We then follow  a heuristic to obtain the \emph{gold attributes} acting as the
referring expression for a given $\langle$referent, context$\rangle$ pair,  by
selecting words that were produced to describe the intended referent but not
the context.  The rationale for this decision is that a \emph{necessary}
condition for achieving successful reference is that REs
accurately distinguish the intended referent from any other object in the
context~\cite{Dale:Haddock:1991}.  For maximizing the quality of the generated
gold attributes, \begin{inparaenum}[(i)] \item we disregarded any distractor
context whose bounding box overlapped significantly with the referent's
bounding box and \item  we disregarded distractor contexts that had full
attribute overlap with the referent, thus resulting in a null referring
expression for the $\langle$referent, context$\rangle$ pair.  \end{inparaenum}

\paragraph{Objects} To control for the complexity of the visual scenes and
attributes while maintaining real images, we created a simpler dataset in which
referent and context are always different objects. For a list of 100 concrete
objects ranging across different categories (e.g., \emph{animal},
\emph{furniture} etc), we synthesized $\langle$referent, context$\rangle$ pairs
by taking all possible combinations of objects and, for each object, sampling an
image from the respective ImageNet~\cite{Deng:etal:2009} object label entry.
The RE/gold attributes for a given pair is then straightforwardly obtained by
using the object name of the referent. 

\paragraph{Shapes} Finally, we introduce a third dataset which controls for the
complexity that real images have,  while allowing referent and context to
differ in a diverse number of attributes, exactly like in real scenes.
Following~\newcite{Andreas:etal:2016}, we created a geometric shapes dataset
consisting of images that contain a single object.  We generate such
single-object images by varying the values of 6 types of attributes and follow
a similar approach as in the \textbf{ReferIt} dataset to annotate
$\langle$referent, context$\rangle$ pairs with gold attributes (i.e., by taking
the difference of the attributes in the referent and context). \footnote{The 18
attributes grouped by type: shape=$\langle$triangle, square, circle$\rangle$,
border color=$\langle$fuchsia, indigo, crimson, cyan, black, limegreen,
brown, gray$\rangle$, horizontal position=$\langle$up, down$\rangle$, vertical
position=$\langle$right, left$\rangle$, shape size=$\langle$small, medium,
big$\rangle$}

\subsection{Agent Players}

\paragraph{Agent A1 (Referring Expression Generation)} Agent A1 is performing a
task analogous to \emph{referring expression generation}.  Unlike traditional
REG research~\cite{Dale:Reiter:1995,Mitchell:etal:2010,Kazemzadeh:etal:2014}
that produces phrases or words as RE, in our current framework, agent A1 learns
to predict a single attribute that discriminates the referent from the context.
Note however that attribute meaning is not pre-defined. Instead, it
emerges ``on-demand'' via their usage in the referential game. In particular,
ideally agent A1 will learn to associate the attributes to systematic
configurations of lower-level perceptual features present in the images. In the
current experiments, the words are simply represented by numerical indices, but
it would of course be trivial to associate such indices with phonetic strings.

\begin{figure}[t]
\center
\includegraphics[scale=0.60]{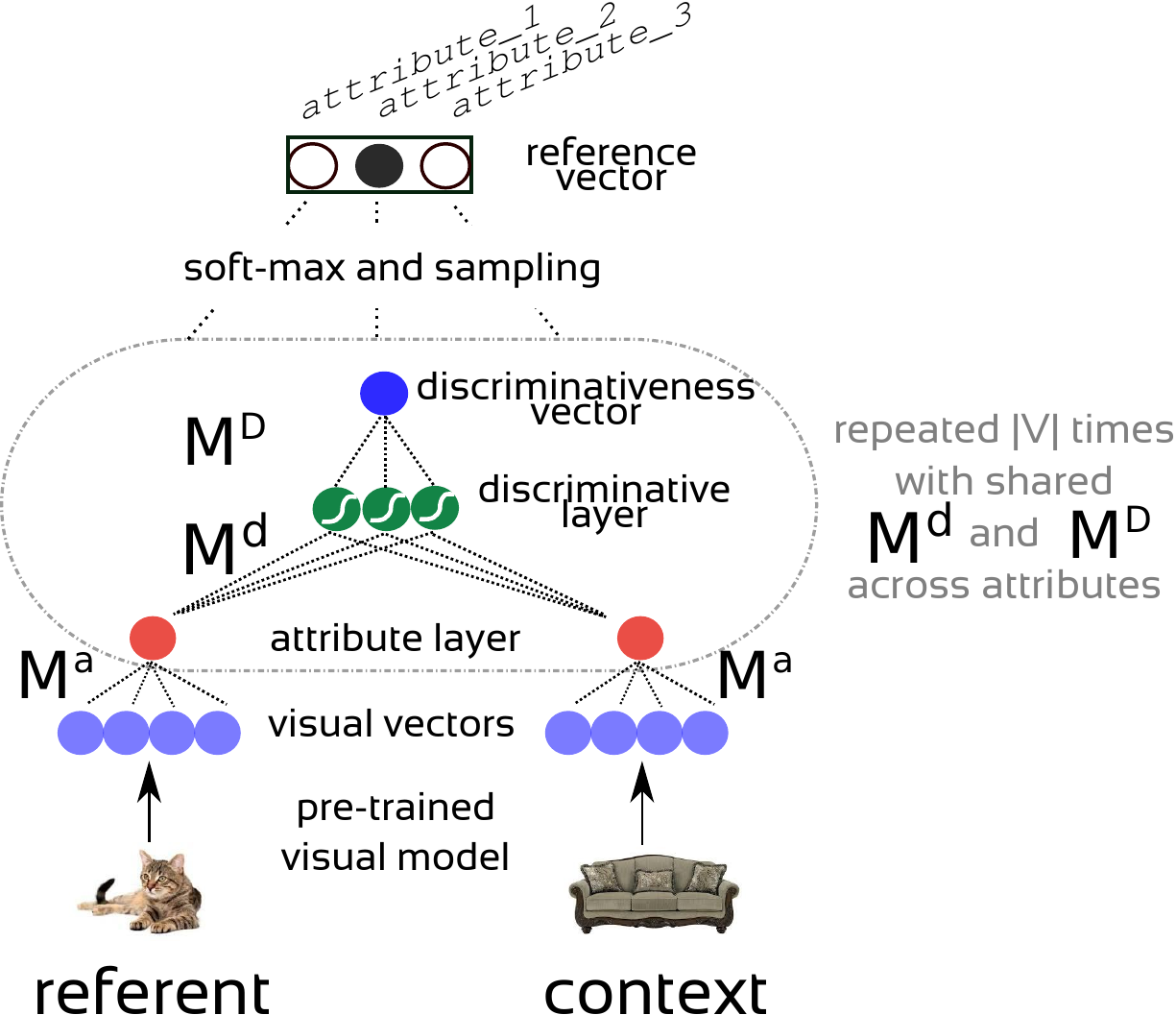}
\caption{Neural network of player A1. In this particular game, A1 produces the reference vector 
that activates \url{attribute_2} as the RE, which is going to be passed over to A1.}
\label{fig:agent1}
\end{figure}

Figure ~\ref{fig:agent1} illustrates the network architecture of A1.
The model is presented with the two images that constitute a visual
scene.  We assume that agents are already equipped with a pre-trained
visual system that converts the raw pixel input of the referent and
the context to higher-level visual vectors $\mathbf{v}$ (i.e., ConvNet
fc7 layers). For games using the \textbf{ReferIt} and \textbf{Objects}
dataset we used the pre-trained
VGG-network~\cite{Simonyan:Zisserman:2014}. For the \textbf{Shapes},
given that the nature of the images are different from the usual
ImageNet data used train these networks, we trained our own
model. Specifically, we trained a smaller network, i.e.,
AlexNet~\cite{Krizhevsky:etal:2012} on predicting bundles of two
attributes. For both models, we use the second-to-last fully connected
layer to represent images with 4096-D vectors.  These visual vectors
are mapped into \emph{attribute} vectors of dimensionality $|V|$
(cardinality of all available attributes), with weights
$\mathbf{M}^\mathbf{a}\in\mathbf{R}^{4096\times{}|V|}$ shared across
the two objects.  Intuitively, this layer learns which attributes are
active for specific objects.

\emph{Pairwise} interactions between attribute vectors of referent
and context are captured in the \emph{discriminative} layer. This layer, processes,
for each attribute, the two units, one for each object, and by applying a
linear transformation with weights $\mathbf{M}^\mathbf{d}\in\mathbf{R}^{2\times{}h}$, 
followed by a sigmoid activation function, finally derives a single value by another linear
transformation with weights $\mathbf{M}^\mathbf{D}\in\mathbf{R}^{h\times{}1}$,
producing $d_v$ which encodes the degree of discriminativeness of attribute $v$
for the specific referent.   
The same process with the same shared weights
$\mathbf{M}^\mathbf{d}$ and $\mathbf{M}^\mathbf{D},$ across attributes is
applied to all attributes $v\in V$, to derive the estimated discriminativeness
vector $\mathbf{d}$. Finally, the discriminativeness vector is converted to a
probability distribution, from which the player samples one  attribute
$a$ acting as the referring expression. This attribute is encoded in the \emph{reference vector}
with one-hot like representation, and is passed over to A2. The learnable parameters of A1 are
$\theta_{A1} = \langle \mathbf{M}^\mathbf{a}, \mathbf{M}^\mathbf{d},
\mathbf{M}^\mathbf{D}\rangle$.

A1 does not receive supervised data regarding the attributes that are
active in pictures, nor  regarding which attributes should be used to refer to
the referent.  The only supervision regarding the ``goodness'' of attributes
for the given $\langle$referent, context$\rangle$ pair comes from the success
of the interaction between the agents while playing the referential game.

\paragraph{Agent A2 (Reference  Resolution)} For the purposes of the game, A2 
 needs to perform a task similar to reference resolution. Given the same
visual input as A1 (we assume that the agents share the same visual system)
and the produced attribute $a$, A2 has to choose which of the 2
objects in the scene is the intended referent.

\begin{figure}[h]
\center
\includegraphics[scale=0.60]{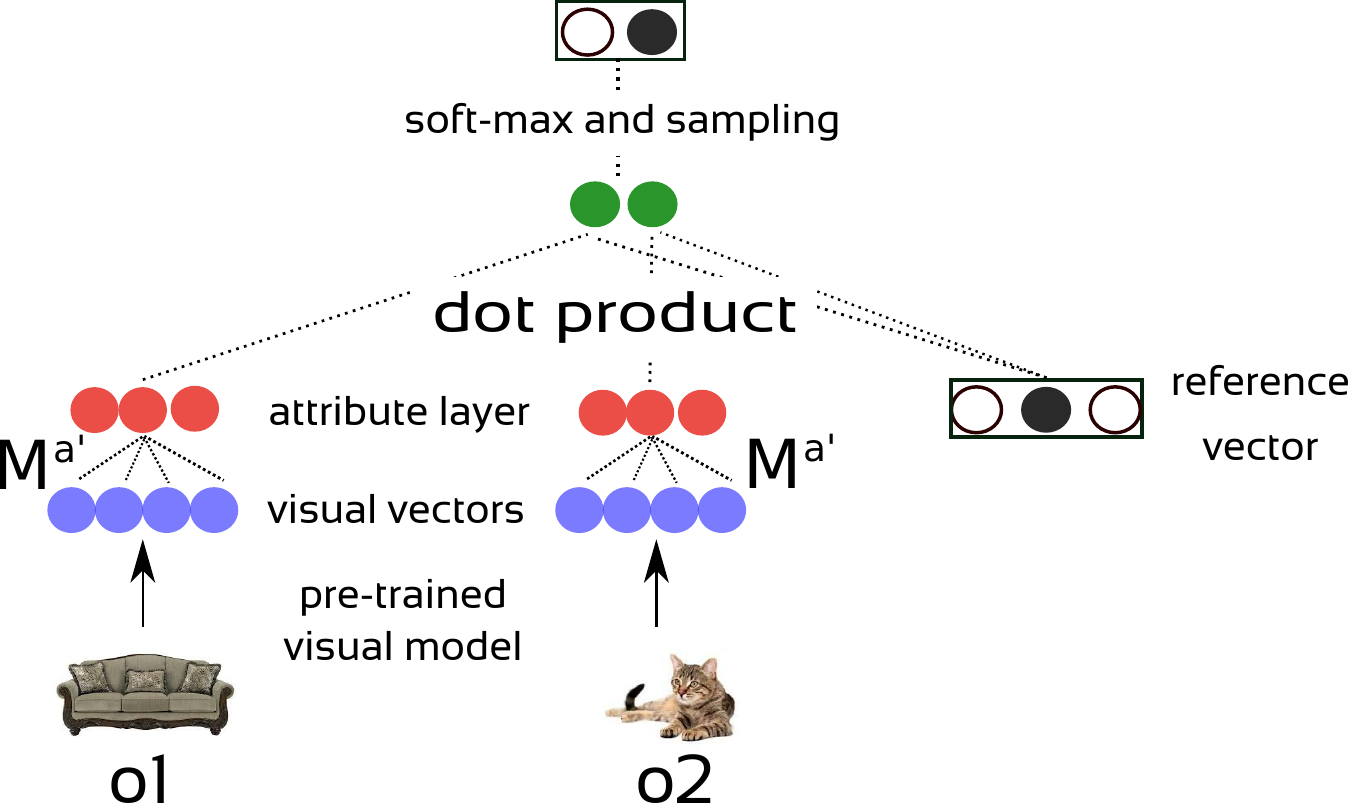}
\caption{Network of A2. Given the two objects and the reference vector encoding the predicted attribute 
produced by A1,  she correctly predicts that the referent is the second object.}
\label{fig:agent2}
\end{figure}

Following this reasoning, we design a simple implementation of A2 depicted in
Figure~\ref{fig:agent2}. A2 is presented with the two objects $o_1$ and $o_2$,
without knowing which is the referent, and embeds them into  an attribute space
using  weights shared across the two objects
$\mathbf{M}^\mathbf{a'}\in\mathbf{R}^{4096\times{}|V|}$.\footnote{While we
could tie  $\mathbf{M}^\mathbf{a'}$ and $\mathbf{M}^\mathbf{a}$, here we do
not enforce any such constraint, essentially allowing the agents to develop
their own ``visual'' understanding.} Note, that as is in the case of A1, A2
will receive no direct image-attribute supervision. The resulting attribute
vectors encode how active the attributes are across the two objects.  Following
that, A2 computes the dot product similarities between the reference vector (i.e., the
one-hot representation of the selected attribute $a$)  and the attribute vectors
of the objects. Intuitively, the reference vector encodes which attribute
characterizes the referent and as a result, the dot similarity will be high if
the attribute $a$ is very active in the attribute vector. These two dot
similarities are converted to a probability distribution $p(o|o_1, o_2, a)$ over
the two image indices, and one index is sampled indicating which of the two
objects is the chosen referent. The learnable parameters of A2 are just $\theta_{A2}=\langle\mathbf{M}^\mathbf{a'}\rangle$

\section{Experiments}

\begin{figure*}[tb]
\includegraphics[scale=0.29]{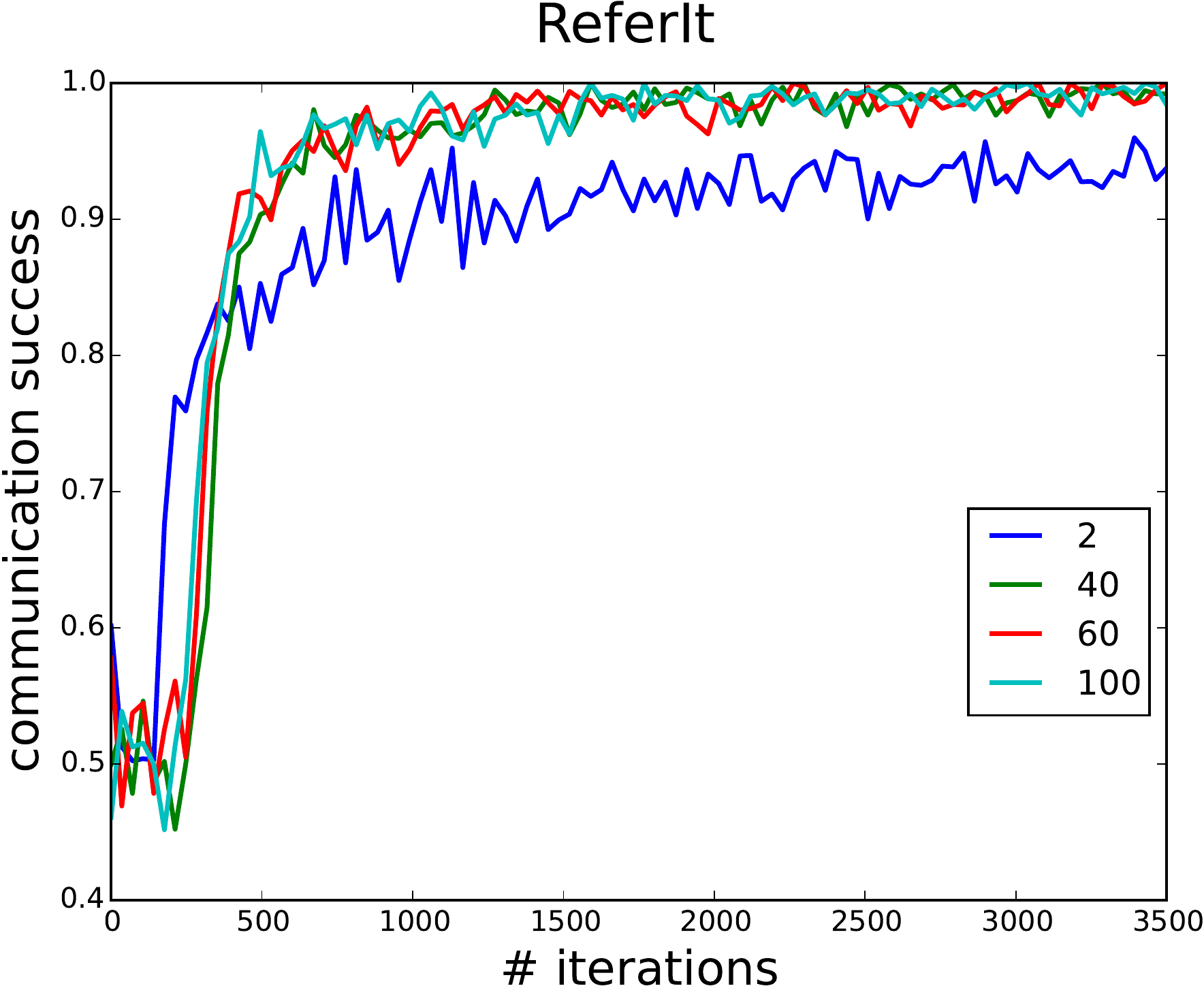}
\includegraphics[scale=0.29]{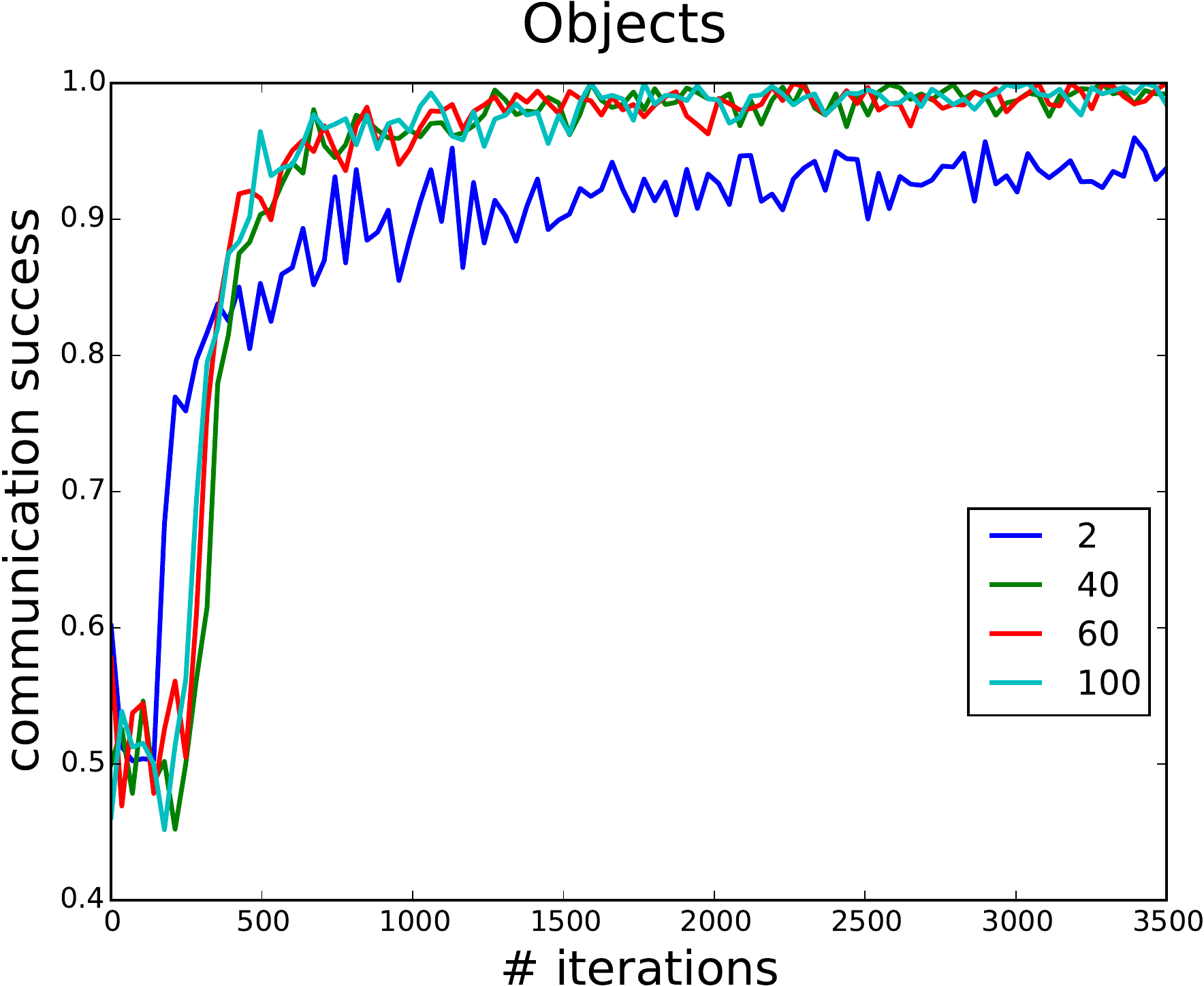}
\includegraphics[scale=0.29]{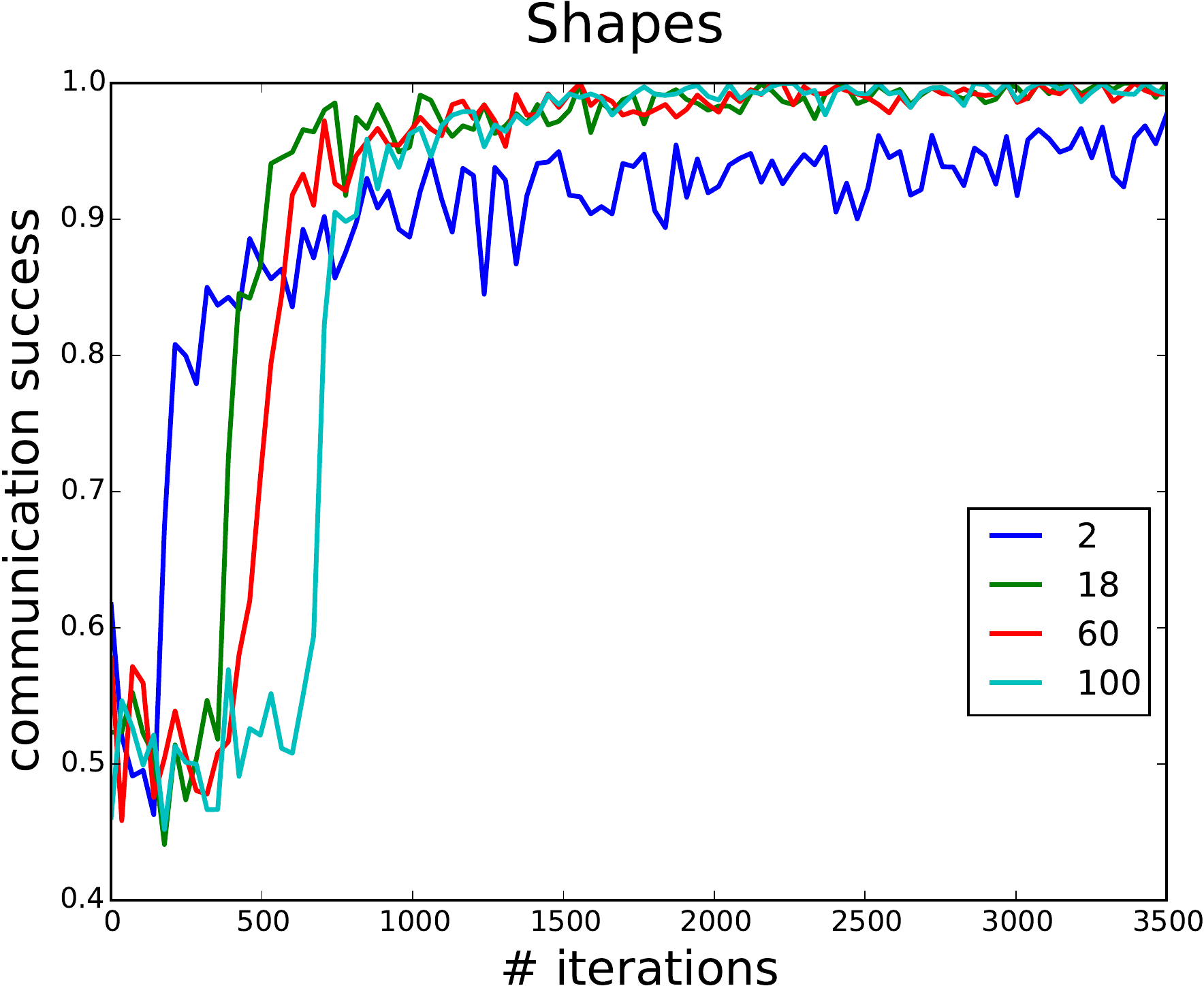}
\caption{Communication success across different datasets over training
  epochs.  For each dataset, we vary the number of attributes
  available in the vocabulary of the agents. \emph{Note: Best viewed
    in color.}}
\label{fig:results_main}
\end{figure*}

\subsection{General Training Details} The parameters of the 2 agents,
$\theta=\langle \theta_{A1}, \theta_{A2}\rangle$, are learned jointly
while playing the game. The only supervision used is communication
success, i.e., whether the agents agreed on the referent. This setup
can be naturally modeled with Reinforcement Learning
\cite{Sutton:Barto:1998}. Under this framework, the parameters of the
agents implement a \emph{policy}. By executing this policy, the agents
perform \emph{actions}, i.e., A1 picks an attribute and A2 picks an
image. The loss function that the two agents are minimizing is $-
\Expect_{\tilde{o}\sim p(o|o_1,o_2,a)}[R(\tilde{o})]$, where $o_1$ and
$o_2$ are the 2 objects in the visual scene, $p(o|o_1,o_2,a)$ is the
conditional probability over the 2 objects as computed by A2 given the
objects and the attribute produced by A1, and $R$ is the reward
function which returns 1 iff $\tilde{o} =$ referent.  The parameter
updates are done following the Reinforce update
rule~\cite{Williams:1992}.  We do mini-batch updates, with a
batch-size of 32 and train in all datasets for 3.5k
iterations.\footnote{The model-specific hidden size of discriminative
  layer hyperparameter is set to $20$.}

The agents are trained and tested separately within each dataset. At test time,
visual scenes (i.e., combination of referent and contexts images) are novel but
individual images might be familiar. For test and tuning we use 1k visual
scenes, and leave the rest for training (see Table~\ref{fig:games} for exact
numbers).

\subsection{Results} Our pilot experiments aim to ascertain whether
our proposal can result in agents that learn to play the game
correctly.  Moreover, given that the agents start from a clean state
(i.e., they possess no prior semantics other than the relatively
low-level features passed to them through the visual vectors), it is
worth looking into the nature of the induced semantics of the
attributes.  Simply put, is communication-based learning enough to
allow agents to use the attributes in such a way that reflects
high-level understanding of the images?

\paragraph{Can agents learn to develop a communication protocol within the referential game?} Figure \ref{fig:results_main} shows the communication
success performance, i.e., how often the intended referent was guessed
correctly by A2 (chance guessing would lead to 0.5 performance).  Overall,
agents are able to come up with a communication protocol allowing them to solve
the task, but it  takes them approximately 500 iterations (i.e., 16k training
examples) before they start communicating effectively. This is to be expected,
as the agents have no knowledge of the game rules nor of how to
refer to things, and must induce both by observing the rewards they receive.
Moreover, fewer attributes in the vocabulary translates to faster, but not
necessarily better learning. As a sanity check, we restricted the vocabulary to
2 attributes only. In this case, the agents also came close to solving the task
(and that was done faster than in the other cases), although without
approaching 100\% performance, a tendency observed across the datasets. At
first glance this might seem suspicious; even if we, as humans, use language
flexibly through polysemy, still it will not be possible to come up with 2
words being able to reliably distinguish all possible combinations of objects!

However, when we closely inspected the way A1 used the 2 attributes (e.g., by
looking into the induced weights $\mathbf{M}^\mathbf{a}$), it became clear that
the agent was in a sense ``cheating''. Instead of communicating about
high-level semantics, the agents agreed to exploit the attributes to
communicate about low-level embedding properties  of $\langle$referent, context
$\rangle$ pairs (e.g., pick attribute 1 if the value in dimension 3 is greater
in the referent than the context etc). While this might seem odd, such
strategies are in fact the best in order to communicate efficiently with only 2
attributes.  This resembles the so-called \emph{conceptual pacts} that humans
form to make conversation more efficient, i.e., mutually agreeing in using
``unconventional'' semantics  to refer to things
\cite{Brennan:Clark:1996}. %
Still, the ``words'' discovered in this way have a very ad-hoc meaning
that will not generalize to any useful task beyond the specifics of
our game, and we would not want the agents to learn them. We thus turn
now to an analysis of the semantic nature of the induced attributes
when the agents have a larger vocabulary available than just 2
attributes, to see if they learn more general meanings, corresponding
to (clusters of) high-level visual properties such as ``red'' or
``cat''.

\begin{figure}[h]
\includegraphics[scale=0.70]{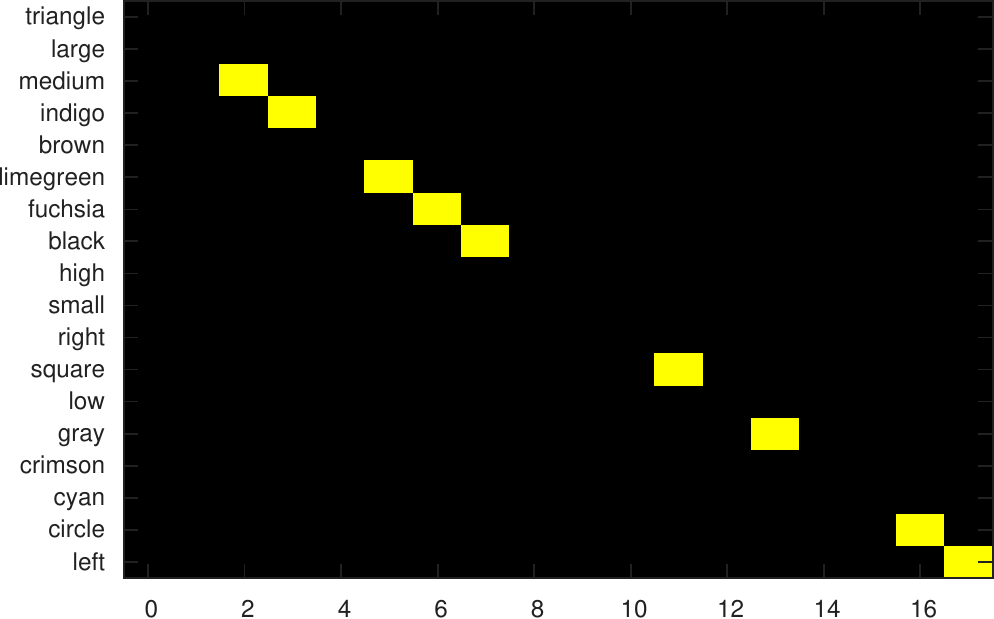}
\caption{Inferred alignment of gold attributes to the induced ones in the \textbf{Shapes dataset}.}
 \label{fig:alignment}
\end{figure}

\paragraph{What is the nature of the induced attributes?} Revealing
the semantics assigned to the induced attributes $a$ is not
trivial. We tested whether the semantics of the induced attributes
align with the semantics of the referring expressions as expressed by
the gold attributes.  We focused on the \textbf{Objects} and
\textbf{Shapes} datasets that have a relatively small number of
attributes (100 and 18 respectively) and trained the agents using as
attribute vocabulary size $|V|$ 100 and 18 respectively.  After the
training, we assign to each induced attribute $a$ the gold attribute
that appears most often in the annotations of the $\langle$referent,
context$\rangle$ pairs for which $a$ was predicted by A1, enforcing a
1-1 mapping (an induced attribute can by paired only with one gold
attribute).  A1 had a tendency  not to make
use of all the available attributes $a$ in the vocabulary, thus either
using attributes in a polysemous way, or  assigning them some semantics
different than the one that the gold attributes encode. As an
example, we plot in Figure~\ref{fig:alignment} the inferred alignment
between induced and gold attributes in the \textbf{Shapes}
dataset. 

Irrespective of the exact interpretation of attributes, meanings
induced within this referential framework should be consistent across
$\langle$referent, context$\rangle$ pairs. For example, if an agent
used the word \emph{red} to refer to the object X in the context of Y,
then \emph{red} cannot be used again to refer to the object Z in the
context of X, since \emph{red} is also a property of the
latter. However, the ``cheating'' approach we reported above for the
2-attribute case, in which the same attribute is used communicate
about whether a low-level feature is higher in the referent or the
context, would not respect this consistency constraint, as X might
have a higher value of the relevant feature when compared to Y but
lower with respect to Z.
 
To capture this, we devised a measure that we termed ``referential
inconsistency'' (\emph{RI}).  Specifically, for each image $i$ we
compute a set $R(i)$ containing all the induced attributes that were
activated when $i$ was in the referent slot and $C(i)$ in the context
slot.  Then, the referential inconsistency $RI$ of an attribute $a$ is
computed as $\frac{\sum_i \big[{a\in R(i)\cap C(i)}\big]}{\sum_i
  \big[{a\in R(i)\cup C(i)}\big]}$, which counts the number of images
for which $a$ was both in the referent and context sets, normalized by
the times it appears in either. Ideally, this value should be 0, as
this happens when it was never the case that an induced attribute was
activated both when an image was in the referent and in the context
slot.

Table~\ref{tab:RI} reports the proportion of induced active attributes
that have $RI>0$ across datasets (smaller values reflect consistent
use of attributes).  As expected, when communicating with 2 attributes
only, agents do not seem to be using them in a semantically meaningful
way, since referential consistency is violated. However, we ought to
mention that it is also possible is that the agents in this case have
learned to use the attributes in a \emph{relative}
way~\cite{Parikh:Grauman:2011}. Imagine that we pair an image of Mona
Lisa once with an image of a frowning face as the context, and once
with an image of a broadly smiling face.  It would be acceptable then
to use ``the smiling one'' to denote Mona Lisa in the first pair, but it would 
also be possible to refer in this way to the more overtly smiling face in the second case, 
as Mona Lisa is more smiley than the frowning face, but definitely less so than a fully
smiling face! In any case, for all datasets, with 100 attributes
referential consistency is largely respected.

\begin{table}[t]
\begin{center}
\small
\begin{tabular}{ c |c c c }
\backslashbox{\textbf{attributes}}{\textbf{datasets}} & \textbf{ReferIt} & \textbf{Objects} & \textbf{Shapes} \\\hline
2 & 1.0& 1.0 &1.0\\
100 & 0.0& 0.03&0.05\\
\end{tabular}
\caption{Proportion of referentially inconsistent  attributes, i.e., 
$RI(a)>0$, across different datasets and with different attribute vocabulary size $|V|$. Smaller values reflect consistent use.}
\label{tab:RI}
\end{center}
\end{table}

\begin{figure}[h]
\includegraphics[scale=0.70]{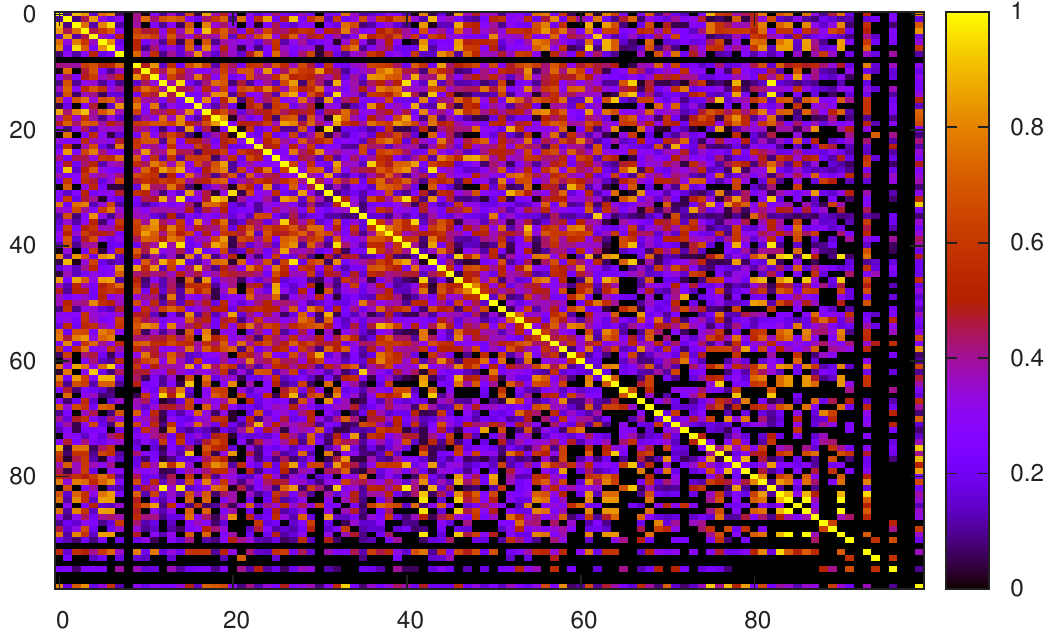}
\includegraphics[scale=0.70]{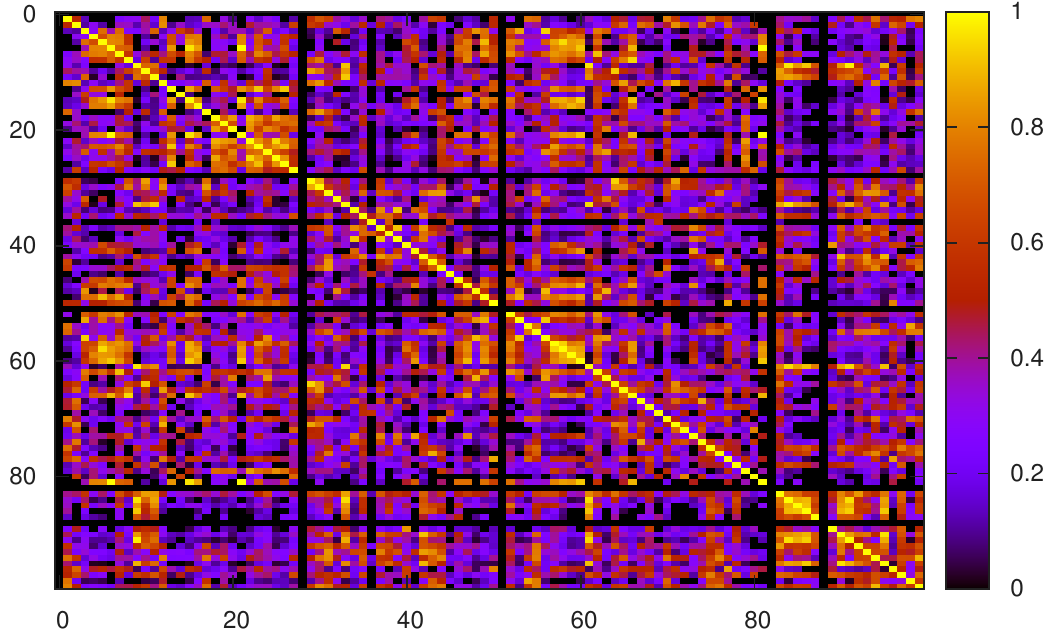}
\caption{Pairwise cosine similarities of gold attributes in induced-attribute vector space in the \textbf{Objects} dataset when ordered randomly (\textbf{up}) 
and according to their category (\textbf{down}).}
 \label{fig:v2_sims} 
\end{figure}

Finally, we consider a third way to assess the degree to which the
induced attributes reflect the intuitive semantic properties of the
images.  Our hypothesis is that, if the induced attributes are used
coherently across visually similar referents, then they should reflect
properties that are typical of the class shared by the referents
(e.g., ``furry'' for mammals).  For this experiment, we focus on the
\textbf{Objects} dataset that is annotated with 100 gold attributes
denoting the objects depicted in the pictures (e.g., \emph{cat},
\emph{dog} etc). For each gold attribute $g$, we construct a vector
that records how often the induced attribute $a$ was used for a
$\langle$referent, context$\rangle$ pair that was annotated with $g$,
essentially representing gold attributes in a vector space with
induced attributes as dimensions. We then compute the pairwise cosine
similarities of the gold attributes in this vector space, plotted in
Figure~\ref{fig:v2_sims} \textbf{up}.  As is, there is no structure in
the similarity matrix.  However, if we organize the rows and columns in the
similarity matrix, as in Figure~\ref{fig:v2_sims} \textbf{down}, so
that objects of the same category cluster together (e.g., the first 2
rows  and columns correspond to \emph{appliances}, the next 4 to \emph{fruits}), 
then a pattern along the diagonal starts to emerge, suggesting
that the induced attributes reflect, at least to same degree,  the similarity that exists between
objects of the same category.

\section{Discussion}
\label{sec:discussion}

We have presented here a proposal for developing intelligent agents with
language capabilities, that breaks away from current passive supervised
regimes. Agents co-exist and are able to interact with each other. In our
proposed framework we do not restrict the number of agents, nor their role in
the games, i.e., we envision a \emph{community} of agents that all interact
with each other having to perform different tasks and taking turns in them, requiring them to either
co-operate or antagonize in \emph{min-max} sort of \emph{zero-sum} games, in
which agents aim at minimizing the opponents gain (e.g., as in the case of the
famous \emph{tic-tac-toe} game).  

In our test case, we considered the most basic act
of communication, i.e., learning to refer to things, and we designed a
``grounded'' co-operative task that takes the form of referential games played
by two agents. 
The first experiments, while encouraging, have revealed that it is
essential to ensure that agents will not ``drift'' into their own
language, but instead they will evolve one that is aligned to our
natural languages.  Thus, inspired by the success of AlphaGo
\cite{Silver:etal:2016}, that combines both passive learning
(experiencing past human games and using a CNN to learn valid moves)
and interactive learning (learning by playing what is the best move
given a particular situation/state), we believe that it is crucial to
combine dynamic interactive learning with static statistical learning
of patterns from association, something that should
ensure the grounding of communication into natural language.

We plan to move along a similar direction, introducing our agents to
multi-tasking. As an example, we could expose A1 to large collections
of texts and train her on language modeling, a task requiring no
manual annotation, from which basic word associations patterns can be
learned. Similarly, A2 could be trained on an image retrieval task,
from which basic concept recognition and naming capabilities can be
acquired, i.e., associating the phonetic string ``cat'' to instances
of cats.  Still, the agents would be trained via playing the game for
producing good referring expressions, which is a task that depends
predominantly on the success of communication (i.e., did our listener
understood what we were referring to?)

\bibliography{../../marco.bib,../../angeliki.bib}
\bibliographystyle{acl2016}

\end{document}